\pdfoutput=1
\PassOptionsToPackage{dvipsnames}{xcolor}

\documentclass[11pt]{article}

\usepackage{ACL2023}

\usepackage{times}
\usepackage{latexsym}

\usepackage[T1]{fontenc}

\usepackage[utf8]{inputenc}

\usepackage[expansion=false]{microtype}

\usepackage{inconsolata}

\usepackage[dvipsnames]{xcolor}
\usepackage{amsmath}
\usepackage{amssymb}
\usepackage{graphicx}
\usepackage{booktabs}
\usepackage{tabularx}
\usepackage{subcaption}
\usepackage{colortbl}
\usepackage{soul}
\usepackage{listings}
\usepackage[export]{adjustbox}
\usepackage{booktabs,array}
\usepackage[title]{appendix}
\usepackage{enumitem}
\usepackage{natbib}

\usepackage{xspace}

\usepackage[normalem]{ulem}

\newcommand{\fullname}{Image Caption Concreteness}
\newcommand{\acronym}{ICC}
\newcommand{\longtae}{semantic-bottleneck autoencoder}
\newcommand{\tae}{SBA}
\newcommand{\longiae}{visual-bottleneck autoencoder}
\newcommand{\iae}{VBA}
\newcommand{\methodname}{\emph{\acronym{}}\xspace}
\newcommand{\methodnamelong}{\emph{\fullname{} (\acronym{})}\xspace}

\definecolor{bubblegum}{rgb}{0.99, 0.76, 0.8}
\definecolor{corn}{rgb}{0.98, 0.93, 0.36}
\definecolor{cream}{rgb}{1.0, 0.99, 0.82}
\definecolor{bluebell}{rgb}{0.64, 0.64, 0.82}
\definecolor{brilliantlavender}{rgb}{0.96, 0.73, 1.0}
\definecolor{brightube}{rgb}{0.82, 0.62, 0.91}
\definecolor{lavenderindigo}{rgb}{0.58, 0.34, 0.92}
\definecolor{lemonchiffon}{rgb}{1.0, 0.98, 0.8}
\definecolor{lightblue}{rgb}{0.68, 0.85, 0.9}
\definecolor{lightgreen}{rgb}{0.67, 0.88, 0.69}
\definecolor{lightred}{rgb}{0.99, 0.5, 0.5}
\definecolor{awesome}{rgb}{1.0, 0.13, 0.32}

\definecolor{lighterred}{rgb}{0.99, 0.7, 0.7}

\newcommand{\hlc}[2][yellow]{{%
    \colorlet{foo}{#1}%
    \sethlcolor{foo}\hl{#2}}%
}

\newcommand{\hlcr}[1]{\hlc[lighterred]{#1}}

\author{Moran Yanuka\:\:\:\:\:\: Morris Alper\:\:\:\:\:\:\: Hadar Averbuch-Elor \:\:\:\:\: Raja Giryes  \\ \\ Tel-Aviv University  \\ \\\url{https://moranyanuka.github.io/icc/}
}

\begin{document}
\title{\methodname{}: Quantifying \fullname{} \\ for Multimodal Dataset Curation}

\maketitle

\begin{abstract}
   Web-scale training on paired text-image data is becoming increasingly central to multimodal learning, but is challenged by the highly noisy nature of datasets in the wild. Standard data filtering approaches succeed in removing mismatched text-image pairs, but permit semantically related but highly abstract or subjective text. These approaches lack the fine-grained ability to isolate the \emph{most concrete} samples that provide the strongest signal for learning in a noisy dataset. In this work, we propose a new metric, \methodnamelong{}, that evaluates caption text without an image reference to measure its concreteness and relevancy for use in multimodal learning. Our unsupervised approach leverages strong foundation models for measuring visual-semantic information loss in multimodal representations. We demonstrate that this strongly correlates with human evaluation of concreteness in both single-word and caption-level texts. Moreover, we show that curation using \methodname{} complements existing approaches: It succeeds in selecting the highest quality samples from multimodal web-scale datasets to allow for efficient training in resource-constrained settings.

\end{abstract}

\section{Introduction}
Pre-training large vision-language models (VLMs) on web-crawled datasets consisting of image-caption pairs has become the standard practice in achieving state-of-the-art results in vision-and-language tasks such as image captioning and multimodal representation learning.
However, raw web data are often noisy and contain many low-quality samples, which impair VLMs' learning in terms of quality and efficiency~\cite{li2022blip,schuhmann2022laion,radenovic2023filtering}.
While various factors impact data quality, we focus on \emph{semantic} noise, characterized by analyzing the meaning of data items rather than, e.g., identifying low resolution images or quantifying token repetitions. 

\begin{figure}[t]
\centering
\setlength{\tabcolsep}{1.3mm}
\hspace*{-0.35cm}\begin{tabular}{>{\centering\arraybackslash}p{0.04\linewidth}  p{0.295\linewidth} p{0.295\linewidth} p{0.295\linewidth} p{0.001\linewidth}}

& 
\includegraphics[trim={0cm 0cm 0cm 0cm},clip, height=2.3cm]{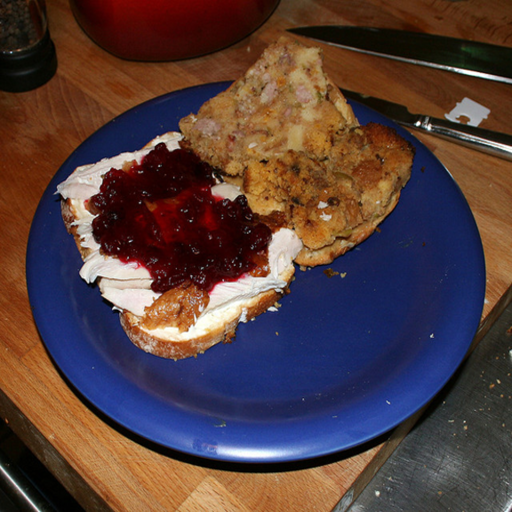} & 
 \includegraphics[trim={0cm 0cm 0cm 0cm},clip,height=2.3cm]{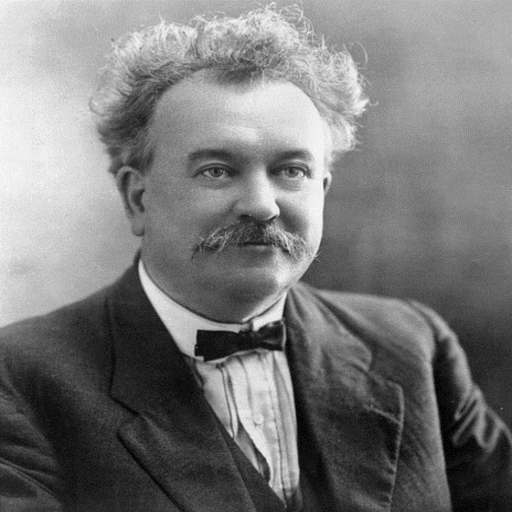} &
\includegraphics[trim={0cm 0cm 0cm 0cm},clip,height=2.3cm]{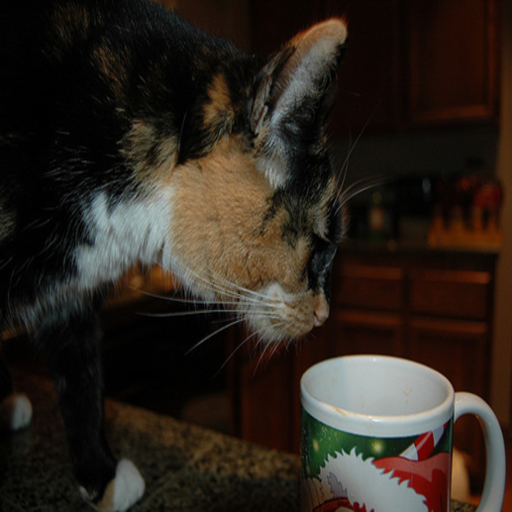} &  \\
\raisebox{-0.025\textwidth}{\centering\large$\uparrow$} & 
\footnotesize{\emph{A sandwich sits on a small blue plate (0.96)}} & 
\footnotesize{\emph{Curly-haired man with a mustache in a vintage photo (1.0)}} & 

 \footnotesize{\emph{A cat standing on a counter looking at a coffee cup (1.0)}} & \\
    \raisebox{-0.018\textwidth}{\centering\large$\downarrow$}&
\footnotesize{\emph{It does not look like something I would eat (0.02)}} & 
\footnotesize{\emph{Talk about a bad hair day, his is frightful (0.01)}}
& 
\footnotesize{\emph{I cant see this image it is too dark (0.02)}} & \\
\end{tabular}
\caption{Given an image caption, \methodname{} measures its visual concreteness. We show samples from MS-COCO~\cite{lin2014microsoft} illustrating captions  annotated by different annotators with low ($\downarrow$) and high ($\uparrow$) \methodname{} scores. As seen above, our method successfully differentiates between concrete and abstract or subjective captions, even for high-quality datasets such as MS-COCO. This is done by quantifying visual-semantic consistency using multimodal foundation models.}
\label{fig:coco}
\vspace{-0.15in}
\end{figure}

\begin{figure}
        \centering %
        \begin{minipage}{0.45\columnwidth}
            \begin{subfigure}[b]{\columnwidth}
                \centering
                \includegraphics[height=50px]{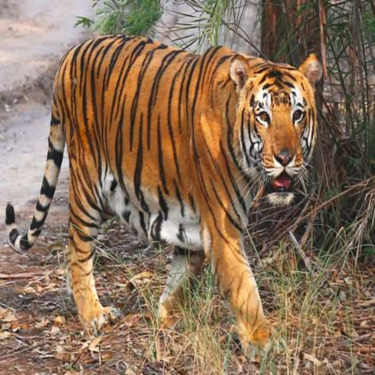}
                \caption{\emph{poaching still remains the biggest threat to tigers}}\label{fig:tiger}
            \end{subfigure}
            
            \vspace{0.5em}
            
            \begin{subfigure}[b]{\columnwidth}
                \centering
                \includegraphics[height=50px]{figures/teaser/clip_teaser_resized/COCO_val2014_000000155897.png}
                \caption{\emph{Wheat bread is always the healthy choice for lunchtime}}\label{fig:bread}
            \end{subfigure}
        \end{minipage}%
        \hfill%
        \begin{minipage}{0.45\columnwidth}
            \begin{subfigure}[b]{\columnwidth}
                \centering
                \includegraphics[height=50px]{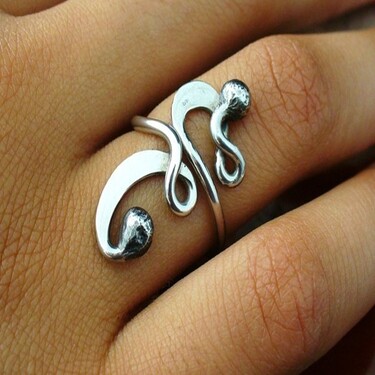}
                \caption{\emph{want a ring like this!}}\label{fig:ring}
            \end{subfigure}
            
            \vspace{0.5em}
            
            \begin{subfigure}[b]{\textwidth}
                \centering
                \includegraphics[height=50px]{figures/teaser/clip_teaser_resized/COCO_val2014_000000255135.png}
                \caption{\emph{Someone did not observe the stop sign and now it is knocked over}}\label{fig:stop_sign}
            \end{subfigure}
        \end{minipage}
        \caption{\textbf{Examples with high CLIP similarity and low \methodname{}.} We show examples from Conceptual Captions dataset (a) and (c), and COCO dataset, (b) and (d). While these captions are semantically related to the images, they are abstract or contain subjective non-visual information that, unlike \methodname{}, CLIP fails to detect.
        } 
\label{fig:teaser}
\end{figure}

Existing datasets are commonly filtered using VLMs such as CLIP~\cite{radford2021learning} to identify image-text semantic misalignments~\cite{sharma2018conceptual,schuhmann2022laion}, i.e. captions irrelevant to their images; using rule-based proxies such as measuring the complexity of captions via semantic parsing~\cite{radenovic2023filtering}; or removing images that contain text that overlaps with the caption~\cite{maini2023t}.
However, these approaches fail to identify captions that are highly abstract and may contain subjective, non-visual information, despite being semantically aligned with the image and having a sufficiently complex grammar.
Figure \ref{fig:coco} shows examples of such image-caption pairs. A caption such as \emph{``It does not look like something I would want to eat''} is semantically related to the image, yielding high CLIP similarity, but contains subjective details which provide a confounding signal when training VLMs (See also Figure \ref{fig:teaser}).  
A model trained to generate such captions from images may learn to hallucinate details, e.g., liking a certain type of food in our example, which are not visually grounded and are highly subjective. Similarly, such image-caption pairs provide a weaker signal for representation learning than images with visually concrete captions (e.g. \emph{``A sandwich sits on a small blue plate''}), which may impede the learning process -- particularly in a resource-restricted setting where data or compute is limited.

Thus, we suggest filtering image captions by their \emph{visual concreteness}, referring to the extent to which a text describes visual aspects of a scene in a manner that can be vividly imagined~\cite{schwanenflugel2013abstract,hessel2018concreteness}\footnote{Some works have treated this as roughly synonymous with \emph{imageability} (visual association), while others use \emph{concreteness} to refer more generally to association with sensory experiences of all types~\cite{richardson1975concreteness,khanna2021well}. Our work focuses on the visual modality.}. This contrasts with abstract text, which may correspond to many possible visual interpretations or include subjective information. We show that this new dimension of textual quality enables selecting image-caption pairs that provide a strong supervision signal for vision-and-language tasks, particularly in resource-constrained settings where training directly on noisy web-scale multimodal data fails to converge to a satisfactory solution in a limited number of iterations.

We propose the \methodnamelong{} metric for quantifying the visual concreteness of image captions calculated from text alone, i.e., without an image reference.
We measure concreteness using unsupervised autoencoding pipelines with visual-semantic information bottlenecks.
Specifically, we use a visual-bottleneck autoencoder that leverages text-to-image generative models' competence and a semantic-bottleneck autoencoder that identifies how well a large language model (LLM) recovers the input caption from its semantic CLIP embedding. As these models require costly inference through large generative models, they cannot feasibly run on a large scale; therefore, our \methodname{} metric is distilled from these pipelines, enabling fast, computationally-efficient inference.

In our experiments, we demonstrate that when dealing with limited training iterations, employing \methodname{} for filtering multimodal datasets leads to enhanced performance in image captioning and representation learning. Moreover, our results indicate a strong correlation between \methodname{} and both single-word concreteness and caption text scores.

Stated explicitly, our contributions are as follows: (1) We propose the \methodname{} metric distilled from foundation VLM models with a novel combination of unsupervised autoencoding pipelines; (2) we show that \methodname{} highly correlates to human concreteness judgements of caption texts; (3) we demonstrate that \methodname{} succeeds in selecting a core of samples from web-scale image-caption datasets for vision-and-language tasks, with superior downstream performance to existing filtering methods.

\begin{figure*}[t]
    \centering
    \includegraphics[width=1.0\linewidth]{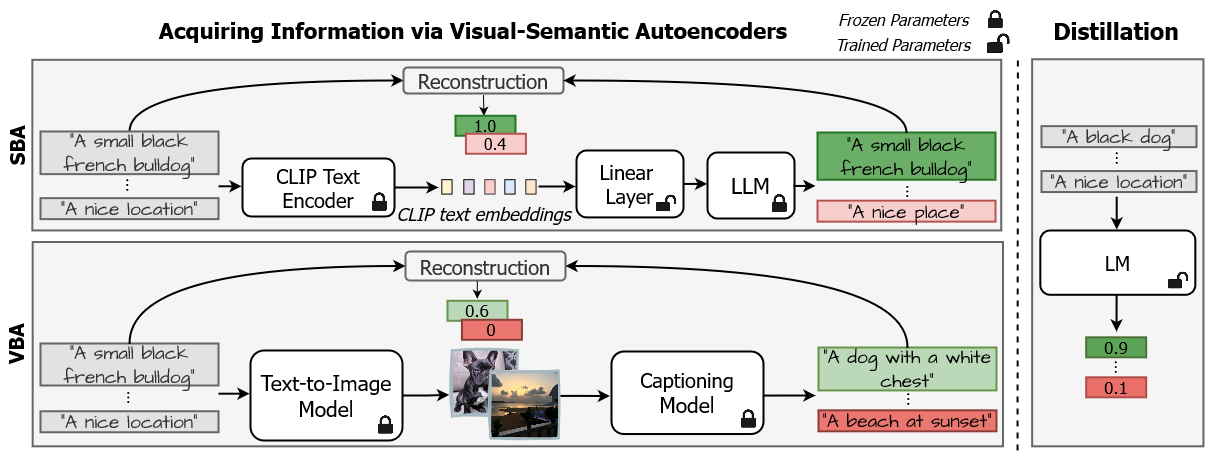}
    \caption{\textbf{\methodname{} pipeline for predicting visual concreteness of image captions.} We first acquire training data using a semantic-bottleneck autoencoder (\tae{}, top left) and an visual-bottleneck autoencoder (\iae{}, bottom left). 
    We then distill a weighted combination of their reconstruction scores into a smaller language model (LM, right), which learns to produce \methodname{} scores for new text. We visualize reconstruction scores for highly concrete (``\emph{A black dog}'') and highly abstract (``\emph{A nice location}'') text. 
    High and low scores are colored in green and red, respectively. %
    Our final score, which combines the two pipelines, yields more accurate concreteness predictions than each of them.} 
    \label{fig:icq_autoencoder}
\end{figure*}

\section{Method}
\label{sec:method}
Given an image caption (of an \emph{unseen} image), we aim to predict its degree of visual concreteness. Our underlying assumption is that more visually concrete text can be mapped to a visual representation with less information loss. Conversely, we expect that visually abstract or subjective text cannot be converted to or from a visual representation without significant information loss, since it does not clearly describe a well-defined image.

As an example, consider the text \emph{``Wheat bread is always the healthy choice for lunchtime''} in Figure \ref{fig:bread}. The notion of wheat bread being a healthy choice is inherently non-visual and is unlikely to be directly depicted in an image. Therefore, this information is likely to be lost in an autoencoding process that includes an image as the bottleneck, when the encoded image is decoded back to the textual modality.

We model this effect with multimodal autoencoders~\cite{kamath2023text,yang2023multicapclip}. In our setting, we use multiple autoencoder components that convert text to and from visual-semantic representations using foundation VLMs, and quantify the information loss of this process as a proxy for visual concreteness. While these autoencoders provide a strong signal, they are composed of slow, computationally-intensive large generative models making inference infeasible on a large scale. Therefore, we distill their scores into a small model which allows for an efficient calculation of the \methodname{} scores.

We proceed to describe our proposed \longiae{}  and \longtae{} components, and their distillation into the final \methodname{} metric. See Figure \ref{fig:icq_autoencoder} for an overview of our full pipeline.

\vspace{1em}
\noindent \textbf{Visual-Bottleneck Autoencoder (\iae{}).} Since a caption represents an image, we construct the \iae{} by using an image as an intermediate representation via which textual information passes.
In particular, we concatenate a frozen text-to-image model (\texttt{Stable Diffusion 2}, \citealp{ramesh2022hierarchical}) and a frozen captioning model (\texttt{BLIP-2}, \citealp{li2023blip}) as shown in Figure \ref{fig:icq_autoencoder} (bottom left). This autoencoding pipeline measures text concreteness by encoding and decoding a caption, followed by measuring semantic fidelity in reconstruction using BERTScore (F1)~\cite{zhang2019bertscore}. We note that this pipeline contains \emph{no trained parameters} as it concatenates pretrained, frozen models.

While the \iae{} pipeline is a simple and intuitive way of enforcing a visual bottleneck, it may sometimes produce sub-optimal reconstructions even for highly visual texts due to its inherently lossy nature. For example, the caption \emph{``a small black french bulldog''} in Figure \ref{fig:icq_autoencoder} may be reconstructed by the \iae{} from the generated image to \emph{``a dog with a white chest"}, which is relatively semantically far from the original caption and thus results in a relatively low reconstruction score of 0.6 for a concrete caption. This stems from the dense information content of generated images, which may contain details (such as the dog's white chest) which were not mentioned explicitly in the original caption, and from the tendency of the captioning decoder to focus on different details than those used to generate the image.
To alleviate this issue, we proceed to propose a complementary method using a stronger prior on caption semantics.

\vspace{1em}
\noindent \textbf{Semantic-bottleneck Autoencoder (\tae{}).}
Motivated by findings that CLIP embeddings encode visual information in text and particularly concreteness~\cite{alper2023bert}, we construct an autoencoding pipeline with CLIP text embeddings as a semantic information bottleneck, as shown in Figure \ref{fig:icq_autoencoder} (top left). We extract visual information from the CLIP text embedding space by utilizing a frozen LLM (\texttt{Llama-2-7b}, \citealp{touvron2023llama}), by training a linear layer that converts CLIP text encoder's output to inputs for the LLM.
The training objective aims at reconstructing the input captions via a token-wise cross-entropy objective. 
By keeping the encoder backbone (CLIP) frozen, this introduces an information bottleneck preventing faithful reconstruction of abstract texts.

After training the \tae{} over image--caption pairs, we use it for measuring text concreteness by encoding and decoding the text followed by measuring reconstruction fidelity. To measure preservation of fine-grained textual details, we quantify this fidelity via per-character edit distance~\cite{levenshtein1966binary}, standardized by caption length, as detailed in Appendix \ref{sec:caption_len}. 

This pipeline generally succeeds in reconstructing highly concrete text (such as \emph{``A small black french bulldog''} shown in the top left part of Figure \ref{fig:icq_autoencoder}). However, the strong textual prior of the \tae{} may also leak information about
abstract and subjective captions as well (e.g. the abstract caption ``\emph{A nice location}'' yields a relatively high reconstruction score of 0.4), limiting its correlation with visual concreteness.
Overall, the \tae{} and \iae{} provide complementary scores, where each correlates more strongly to visual concreteness in different cases. Therefore, they perform most strongly when combined together, as we explicitly verify in our ablations in Section \ref{sec:ablations}. We also show qualitative examples in figures \ref{fig:vba_sba_conc} and \ref{fig:vba_sba_abs} in the appendix. 

\vspace{1em}
\noindent \textbf{\methodname Distillation.} 
Using the aforementioned pipelines to quantify the concreteness at scale is not feasible, as this requires running large models (e.g., diffusion models, LLMs) with billions of parameters for many forward passes per instance (up to dozens of forward passes for the diffusion models inference and for the LLM and captioning model decoding). This requires more than 1,000 GPU hours for a dataset of 1M samples. Therefore, we assemble \tae{} and \iae{} reconstruction scores over a relatively small collection of image-caption pairs and distill their aggregated values into our final \methodname{} score. This enables efficient inference that can easily run on a large scale, with over a hundred times faster inference time and much less compute required. Specifically, we train a small text encoder model~\cite{liu2019roberta} to predict a logit-linear combination of the \tae{} and \iae{} scores, computed as described in the appendix.

\vspace{1em}
\noindent \textbf{Implementation Details of \methodname{} Construction.}  For the construction of our \methodname{} score, we use a subset of CC3M~\cite{sharma2018conceptual} composed of 595K image-caption pairs, introduced by \citet{liu2023visual} and designed to have wider concept coverage. We take a subset of 476K samples for training the linear layer of the \tae{}, and train for 2 epochs with a batch size of 128 and learning rate of 2e-3 with cosine scheduling function. The remaining 118K samples are used for generating reconstruction scores through the \iae{} and the trained \tae{}. For each input caption, we generate five reconstructed captions using beam search (five beams) with the \iae{}'s captioner and the \tae{}'s LLM and then choose the reconstructed caption with the highest similarity to the source caption. By generating the reconstructions and measuring the reconstruction fidelities, we obtain a dataset of 118K captions and corresponding reconstruction scores. We standardize by caption length to disentangle the dependency of the reconstruction scores to the caption length (i.e., forcing the same distribution of scores for all caption lengths), as described in the appendix. We train a small language model (\texttt{DistillRoberta-Base}) to predict the combined scores on these samples with a Mean Squared Error objective. This final distilled model is used for generating the \methodname{} scores.

\section{Results}

We turn to show \methodname{'s benefit in data curation for downstream tasks (Section \ref{sec:curation-eval}), followed by its correlation to human judgement (Section \ref{sec:concretness-eval}).

\subsection{VLM Dataset Curation}
\label{sec:curation-eval}
\vspace{0.8em}
\textbf{Experimental Settings.}
We investigate the effect of \methodname{} and other filtering methods for curating a core of high-quality image-caption pairs from large multimodal datasets, comparing their effects on downstream task performance -- both discriminative (representation learning) and generative (image captioning).
 We follow similar settings as described in the Datacomp~\cite{gadre2023datacomp} benchmark's filtering track\footnote{As opposed to the BYOD track which allows for modifying the samples, for instance by using synthetic captions.}, with the following modifications to model the resource-limited setting: given a training dataset comprised of $\mathcal{M}$ samples, the downstream model is constrained to train for exactly $\mathcal{N} \ll \mathcal{M}$ iterations over the filtered subset of the dataset. 
This contrasts with the original Datacomp setting where $\mathcal{N} = \mathcal{M}$, which requires significant compute for a web-scale dataset. Our formulation tests the ability of filtering methods to curate high-quality core subsets of such datasets.
Our initial subset of LAION-400M is composed of $\mathcal{M}=8M$ samples and we fix $\mathcal{N}=2M$ training iterations. To verify the robustness of our method, we measure downstream performance over visually grounded benchmarks across three different sizes of filtering.

We compare to four existing filtering methods -- CLIPScore ~\cite{hessel2021clipscore}, Complexity and Action (CA)~\cite{radenovic2023filtering}\footnote{Using our re-implementation, as there is no publicly available code.}, T-MARS~\cite{maini2023t}, and PACScore \cite{sarto2023positive}. CA is a rule-based filtering method which aims to retain only sufficiently complex captions that also contain an action, based on semantic parsing. T-MARS filters multimodal datasets by removing samples whenever an image includes text that overlaps significantly with the caption. PACScore trains a CLIP-based model with positive-augmented contrastive learning approach, showing improved correlations with human intuition in scoring image-caption pairs.
As opposed to these methods, we focus on filtering according to the concreteness of image captions.

\begin{figure*}[h]
\centering
\begin{adjustbox}{width=\textwidth,center} 
\renewcommand{\arraystretch}{2.0} %
\setlength{\tabcolsep}{1.65mm}
\hspace*{-0.35cm}\begin{tabular}{>{\large\centering\arraybackslash}p{0.065\linewidth}  p{0.18\linewidth}  p{0.165\linewidth} p{0.12\linewidth} p{0.18\linewidth} p{0.17\linewidth} p{0.001\linewidth}}

& 
\includegraphics[trim={2cm 0cm 0cm 0cm},clip, height=2.3cm]{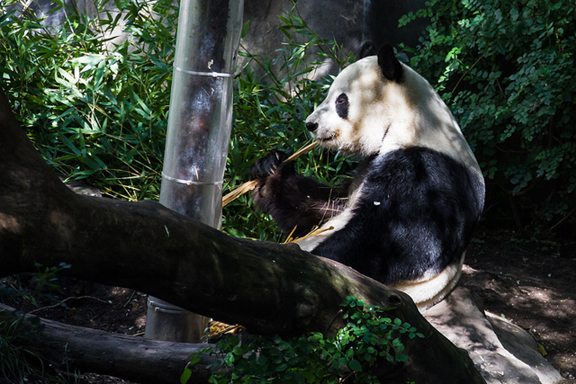} & 
\includegraphics[trim={0cm 0cm 0cm 0cm},clip, height=2.3cm]{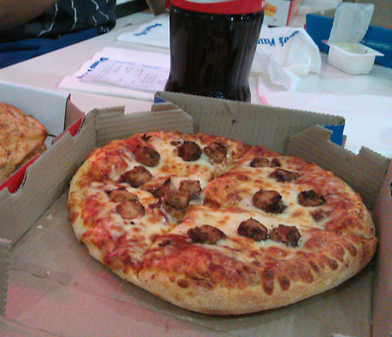} &

\includegraphics[trim={0cm 0cm 1.0cm 0cm},clip,height=2.3cm]{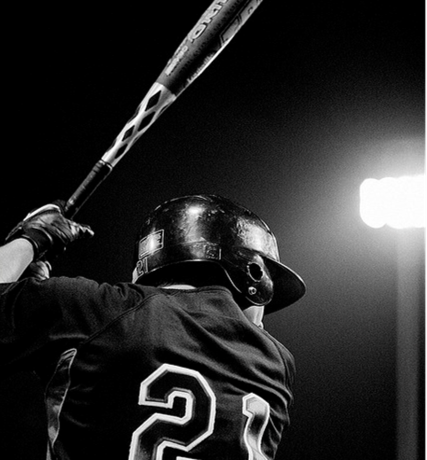} & 
\includegraphics[trim={0.2cm 0cm 0.5cm 0cm},clip, height=2.3cm]{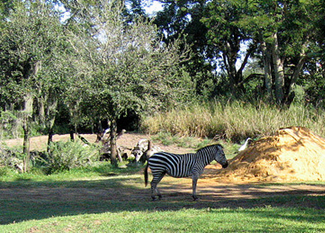} & 
\includegraphics[trim={2cm 0cm 0cm 0cm},clip,height=2.3cm]{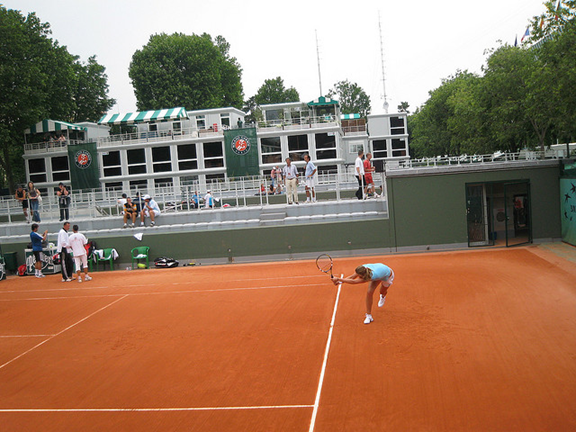} \\

\raisebox{-0.018\textwidth}{CA} &
\raggedright
\footnotesize{\emph{Tiger cubs playing in the rain at the Zoo of the Ozarks in Washington, D.C. on Saturday, Oct. 18}} & 

\footnotesize{\emph{Coffee at the bar.  I love this place!  It's a great way to get away from the hustle and bustle}} &

\footnotesize{\emph{Cleveland Indians vs. Boston Red Sox}} & 
\footnotesize{\emph{Cambodia, the largest of all the African savannahs, is one of the most arid regions in the world.}} & \footnotesize{\emph{Rugby World Cup 2019: The men's singles final takes place at the Ritz-Carlton in London, England, on Saturday,}}  \\

\raisebox{-0.018\textwidth}{TMS} &

\footnotesize{\emph{Catching a lion in the wild is one of the most beautiful things you can do in the wild.}} & 
\footnotesize{\emph{Coffee at the bar. Photo credit: Flickr userfairy.com.au. (via Flickr)}} &

\footnotesize{\emph{Buster Posey hits a two-run home run...}} & 
\footnotesize{\emph{Aerial view of a wild boar in a field in Namibia, South Africa, Africa. (Photo courtesy of the Namibian Wildlife)}} & \footnotesize{\emph{Astonishingly, there was no shortage of competition between the two-school teams at the London 2012 Olympic Games.}}  \\

\raisebox{-0.018\textwidth}{CLIP} &

\footnotesize{\emph{Polar bear cubs pose for a photo with a polar bear in the background. Credit: NASA/JPL-Caltech/UCLA}} & 
\footnotesize{\emph{Pizza Hut Creamy Pizza Sandwich with Bacon, Cheese, and Tomato Sauce}} &

\footnotesize{\emph{Bryce Harper of the Toronto Blue Jays signs autographs for fans prior to the
game}} & 
\footnotesize{\emph{Aerial view of the world's largest crocodile in the Serengeti National Park.}} & \footnotesize{\emph{New York Mets Fanatics Authentic 8"" x 10"" Skateboard Deck}}  \\

\raisebox{-0.018\textwidth}{\methodname} & 
\footnotesize{\emph{Panda eating bamboo}} &
\footnotesize{\emph{A picture of a pizza box full of pizzas}} &

 \footnotesize{\emph{black and white photo of a baseball player}} &  
 \footnotesize{\emph{Zebra at the zoo}} & \footnotesize{\emph{A view of the tennis court from the front.}} \\

\end{tabular}
\end{adjustbox}
\vspace{3pt}
\caption{Qualitative examples of captions generated by captioning models trained on datasets filtered with different filtering methods, over images from MS-COCO test split. CA denotes Complexity and Action filtering, and T-MARS is marked by TMS. As seen above, models trained on \methodname-filtered data generate much more concrete and visually-grounded captions.}
\label{fig:qual_cap}
\end{figure*}

\begin{table*}[h]
  \centering
  \setlength{\tabcolsep}{2.8pt}
    \begin{tabular}{l|c|ccccccc|ccccccc}
    
    \multicolumn{2}{c}{} & \multicolumn{7}{c}{\textbf{MS-COCO}} & \multicolumn{7}{c}{\textbf{NoCaps}}\\
    \midrule
      \textbf{Method} & \textbf{\# Samples} & B@4 & M & R & C &  S & BSc & P  & B@4 & M & R & C &  S & BSc & P \\

        \midrule

        Random
        & 100k
        & 0.9 &	4.7 & 11.2 &	5 &	2.3	 & 0.64 & 0.18 &	1 &	5.1 &	11.9 &	5.6 &	1.6 &	0.68 & 0.11 \\ 

        CLIP
        
        & 100k
        & 1.1 &	5.5 &	11.9 &	2.5 &	2.2 &	0.75 & 0.12 &	1.4 &	5.7 &	12 	& 3 &	1.5 &	0.71  & 0.08\\ 

        CA 
        & 100k
        & 0.9 &	3.7 &	7.3 &	3.2 &	1.6 &	0.27 &	0.20 & 1.6 &	4.4 &	9.4 &	4.1 &	1.2 &	0.33 & 0.11 \\    

        T-MARS
        & 100k
        & 1.2 &	4.6 &	10.6 &	5.6 &	2.3 &	0.53 &	0.20 & 1.3 &	4.9 &	11.6 &	6.3 &	1.7 &	0.61 & 0.11   \\
        
        PACScore
        & 100k
       & 1.9 & 6.8 & 15 & 5.5 & 3 & 0.82 &  0.16 & 2.9 & 7.5 & 16.1  & 7.7 & 2.4 & 0.79 & 0.1 \\

        \methodname
        & 100k
        & \textbf{10.1} & \textbf{15.4} & \textbf{35.4} & \textbf{35.8} & \textbf{10.3} & \textbf{0.9} & \textbf{0.39} & \textbf{12.1} & \textbf{15.8}	& \textbf{35.9}	& \textbf{33.3}	& \textbf{6.4}	& \textbf{0.9} & \textbf{0.2}  \\     

        \midrule
        Random
        
        & 200k
        & 0.9 &	4.2 &	9.8 & 	5 &	2.2 &	0.51 & 0.19  &	1 &	4.8 &	11.2 &	5.9 &	1.7 &	0.6  & 0.11 \\

        CLIP
        & 200k
        & 1.3 &	5.7 &	12.4 &	3.4 &	2.6 &	0.72 &	0.21 & 1.6 &	6 &	12.6 &	3.5 &	1.8 &	0.67  & 0.09  \\     

        CA 
        & 200k
        & 0.5 &	2.8 &	5.7 &	3.7 &	1.3 &	0.18 & 0.20  &	1.2 &	3.4 &	7.2 &	4.1 &	1.1 &	0.24  & 0.12   \\    

        T-MARS
        & 200k & 
        1.1	& 4.6	& 10.7	& 6.5	& 2.4	& 0.5	 & 0.21 & 1.7	& 5.4	& 12.3	& 7.8	& 1.9	& 0.6  & 0.12 \\

        PACScore
        & 200k
       & 2.9 & 7 & 15.5 & 7 & 3.7 & 0.73 & 0.19 & 3.8 & 7.4 & 15.8 & 8.8 & 2.6  & 0.67  & 0.12  \\

        \methodname
        & 200k
        & \textbf{10} & \textbf{15.2} & \textbf{34.6} & \textbf{35.5} & \textbf{10.4} & \textbf{0.9}  & \textbf{0.39} & \textbf{13.1} &	\textbf{15.8} &	\textbf{35.2} &	\textbf{34.3}	& \textbf{6.7} &	\textbf{0.9}  & \textbf{0.21}  \\ 

        \midrule
        Random
        
        & 500k
        & 0.6 &	3.4 &	8 &	4.5 &	1.9 &	0.42 &	  0.2 & 0.9 &	4.2 &	10.1 &	5.5 &	1.5 &	0.55  & 0.12  \\

        CLIP
        & 500k
        & 5.2 &	9.4 &	22 &	15.1 &	5.3 &	0.8 & 0.24 &  5.2	& 8.9  &	21.3 &	12.9 &	3	& 0.8  & 0.13  \\     

        CA 
        & 500k
        & 0.7 &	3.1 &	6 &	3.6 &	1.4 &	0.19  & 0.2  &	2.1 &	4.5 &	9.4 &	5.3 &	1.5 &	0.29  & 0.13   \\   

        T-MARS
        & 500k
       & 0.8 & 3.7 & 8.9 & 5.7 & 2 & 0.42  & 0.21 & 1.2 & 4.7 & 10.8 & 6.5 & 1.7 & 0.65  & 0.12  \\

    PACScore
        & 500k
       & 2.6 & 6.5 & 15 & 8.6 & 3.7 & 0.65 & 0.21 & 3 & 6.9 & 15.4 & 10.4 & 2.6  & 0.65  & 0.13  \\

    \methodname
        & 500k &
         \textbf{8.3} &	\textbf{13.9} &	\textbf{31.4} &	\textbf{30.9} &	\textbf{9.7} &	\textbf{0.89}  & \textbf{0.37} &	\textbf{10}	& \textbf{14.2} &	\textbf{31.3} &	\textbf{28.2} &	\textbf{6} &	\textbf{0.89}  &  \textbf{0.2} \\

        \bottomrule
    \end{tabular}
    
  \caption{\textbf{Captioning results for different filtered dataset sizes}. We perform evaluation of captioning models over MS-COCO and NoCaps datasets trained over different filtering schemes of the LAION-400M dataset, with varying dataset sizes. We compare the performance of \methodname to five filtering baselines. Among these, Random refers to random samples from LAION-400M, CLIP indicates filtering by top CLIPScore, and CA indicates Complexity and Action filtering. B@4, M, R, C, S, BSc and P denote BLEU-4, METEOR, Rouge-L, CIDEr, SPICE, BERTScore and Polos metrics respectively. \# Samples denotes the amount of samples retained after filtering. Best results are in \textbf{bold}.
}
  \label{tab:full_results_captioning}
\end{table*}

\vspace{1em}
\noindent
\textbf{Captioning Models.} 
\label{sec:captionig}
In Table \ref{tab:full_results_captioning} we show quantitative results of applying \methodname{} filtering on top of standard CLIPScore filtering over the subset of LAION-400M for training a captioning model.
The captioning model used is an encoder-decoder architecture with a pretrained Swin~\cite{liu2021swin} vision encoder and GPT-2~\cite{radford2019language} text decoder. We use a batch size of 100, and learning rate of 2e-5 with a cosine scheduler.
We test our approach over two standard captioning benchmarks datasets -- MS-COCO~\cite{lin2014microsoft} and NoCaps ~\cite{agrawal2019nocaps}, across multiple captioning metrics \cite{papineni2002bleu, banerjee2005meteor, lin2004rouge, vedantam2015cider, anderson2016spice, zhang2019bertscore, wada2024polos}.  
As illustrated in the table, filtering with \methodname{} outperforms by a large margin the alternative filtering methods for captioning given a fixed number of desired samples and training iterations. Note that unlike other methods, \methodname{} is directly aligned with the captioning objective, as a captioning model should generate visually-grounded concrete text. This may explain the large gap in performance between \methodname and other filtering baselines. We show qualitative comparison between captioning models trained with different filtering methods in Figure \ref{fig:qual_cap}, exemplifying how filtering with \methodname promotes more concrete and accurate captioning.
 
\begin{table}[h]
  \centering
  \setlength{\tabcolsep}{1.45pt}
    \begin{tabular}{l|c|ccc|ccc}
    
    \multicolumn{2}{c}{}  & \multicolumn{3}{c}{\textbf{COCO}} & \multicolumn{3}
    {c}{\textbf{Flickr}} \\
    \midrule
      \textbf{Filt.} & \textbf{Size} &  R@1  & R@5  & R@10  & R@1  & R@5  & R@10 \\

        \midrule

        Rand.
        & 100k &
        5 & 15.4 & 23.3 & 10.6 & 31.5 & 42.6  \\ 

        CLIP
        
        & 100k
         & 2.1 &	7.5 & 12.4 & 5.7 &	17.1 &	26  \\ 

        CA 
        & 100k
         & 5.2 & 15.8 & 24.1 & 11.3 & 32.2 & 43.8  \\    

        TMS
        & 100k
        & 6.5 & 19.5 & 28.8 & 14.9 & 37.1 & 49.5  \\
        
        PAC 
        & 100k
        & 4.8 & 14 & 21.2 & 9.2 & 24.7 & 35.5 \\

        \methodname
        & 100k  & \textbf{14.4} & \textbf{34.5} & \textbf{45.7} & \textbf{32.6} & \textbf{62.7} & \textbf{73.5}  \\     

        \midrule
        Rand.
        
        & 200k
         & 9.6 &	25.5 &	36.2 &	21.1 &	48.9 &	61.8  \\

        CLIP
        & 200k
         & 6.9	& 10 &	15.8 &	6.9 &	20.9 &	30.9 \\     

        CA 
        & 200k &
        8.8	& 24.4	& 35.1 & 20.8 &	48.6 &	61.2  \\    

        TMS
        & 200k
         & 8.2 &	23 & 32.8 &	17.8 &	43.4 &	56.3  \\
        PAC 
        & 200k
        & 6.5 & 17.7 & 26.3 &	12.9 &	31.1 &	42.9 \\

        \methodname
        & 200k
         & \textbf{15.5} & \textbf{35.8} & \textbf{47.6} & \textbf{33.6} & \textbf{63.2} & \textbf{74.5}  \\ 

        \midrule
        Rand.
        
        & 500k
        & 8 & 22.2 & 32.5 & 17.4 & 42.1 & 55.4  \\

        CLIP
        & 500k
         & 5.3 &	16 &	23.9 &	11.5 &	30 &	42.7 \\     

        CA 
        & 500k
        & 8.2 & 22.6 & 32.4 & 17 & 43.3 & 56.7  \\   

        TMS
        & 500k
        & 10 &	26.3 &	37.2 &	20.3 &	46.8 &	60.5   \\

        PAC 
        & 500k
        & 8.8 & 23.5 & 33.9 & 17.8 & 40.4 &	53 \\

        \methodname
        & 500k
        & \textbf{14.6} & \textbf{34.9} & \textbf{47} & \textbf{30.6} & \textbf{60.9} & \textbf{72.9}   \\

        \bottomrule
    \end{tabular}
    
  \caption{\textbf{Representation learning results over different filtered dataset sizes}. We perform text-to-image retrieval evaluation over MS-COCO and Flickr30K for different filtering schemes of LAION-400M with varying dataset sizes. We compare our performance (\methodname{}) to various filtering baselines: Rand. indicates selecting random samples from LAION-400M, CLIP indicates filtering by top CLIPScore, CA indicates Complexity and Action filtering, TMS indicates filtering with T-MARS and PAC indicates filtering with PACScore. Best results are in \textbf{bold}. 
}
  \label{tab:full_results_clip}
\end{table}

\vspace{1em}
\noindent
\textbf{Image-Text Representation Learning.}
We also perform a representation learning experiment by training a dual text and image encoder model on LAION-400M filtered with different methods. Table \ref{tab:full_results_clip} reports text-to-image retrieval over standard held-out retrieval benchmarks, namely MS-COCO~\cite{lin2014microsoft} and Flickr30K~\cite{plummer2015flickr30k}. The model is initialized from pretrained vision and text encoders (\texttt{ViT-base}, \texttt{BERT-Base})~\cite{dosovitskiy2010image,devlin2018bert}, as suggested by \citet{zhai2022lit}. We use a batch size of 128, learning rate of 2e-5 with a cosine scheduling function. All other filtering methods in the table are identical to the ones in the captioning setting. As illustrated in the table, \methodname{} yields superior performance for this task, showing that our method selects samples which provide better signals for downstream retrieval applications in this setting. 

We note that although prior work has found filtering methods such as CLIPScore~\cite{hessel2021clipscore} to be beneficial~\cite{gadre2023datacomp}, we find that it fails to significantly improve (or even degrades) results in the case of selecting a small core of samples. This accords with previous work showing that applying filtering to LAION-400M with CLIP degrades the performance~\cite{maini2023t} in some of the benchmarks, likely due to high-scoring images containing literal text that overlaps with the caption.

\subsection{Concreteness Correlation} \label{sec:concretness-eval}

\begin{table}[t]
  \centering
  \setlength{\tabcolsep}{3pt}
  \begin{tabular}{lccc|ccc}
      \toprule
     & \multicolumn{3}{c}{\textbf{Word Conc.}} & \multicolumn{3}{c}{\textbf{Caption Conc.}}\\
\midrule
    Method & $\rho$  & $\rho_s$  & $\tau$  & $\rho$  & $\rho_s$  & $\tau$  \\
    \midrule
    CLIP-SP   & 0.6 & 0.62 & 0.44 & 0.34 & 0.33 & 0.25\\
    
    aveCLIP & 0.55 & 0.56 & 0.39 & 0.29 & 0.28 & 0.22 \\

     GPT-3.5 & 0.55 & 0.56 & 0.44 & 0.44 & 0.48 & 0.4 \\
    
   GPT-4o & \textbf{0.78} & \textbf{0.79} & \textbf{0.64} & \underline{0.57} & \underline{0.57} & \underline{0.49} \\
    \textbf{\methodname} & \underline{0.75}
    & \underline{0.75} & \underline{0.55} & \textbf{0.73} & \textbf{0.75} & \textbf{0.6}\\
    \bottomrule
  \end{tabular}
  \caption{\textbf{Concreteness evaluation on single-word and caption-level texts.} Correlation 
  (in absolute value) is measured using %
  Pearson $\rho$, Spearman $\rho_s$, and Kendall $\tau$ coefficients. Best result are in \textbf{bold}, second best are \underline{underlined}.}
  \label{tab:corr}
\end{table}

Table \ref{tab:corr} shows the correlations of different concreteness estimation methods to ground-truth concreteness scores on both single-word and caption-level benchmarks.
We compare \methodname{} to three baselines. The first baseline is zero-shot probing of CLIP through Stroop probing (SP) as proposed by \citet{alper2023bert}. The second baseline is aveCLIP~\cite{wu2023composition}, a learned metric quantifying concreteness at the sentence level, which generates multiple images from a caption and measures the average CLIP-similarity between the text and generated images. Due to its high computational cost, we evaluate it on a random subset of each benchmark (as described in the appendix). Finally, we compare to GPT-3.5-Turbo and GPT-4o~\cite{achiam2023gpt}, used in the zero-shot setting by prompting them to provide concreteness scores. The prompts used are detailed in the appendix.

\vspace{1em}
\noindent
\textbf{Correlation to Word Concreteness.} We first validate our metric by measuring it on the dataset introduced by \citet{hessel2018concreteness}. This consists of 39,954 English unigrams and bigrams coupled with human-labelled concreteness scores on a scale from 1 (abstract) to 5 (concrete), averaged over annotators. To compare with prior work, we only use unigram nouns, totaling 14,562 items.
As seen in Table \ref{tab:corr}, \methodname{} outperforms prior dedicated methods for measuring word concreteness, while performing competitively with the proprietary and much larger GPT-4o.

\vspace{1em}
\noindent
\textbf{Correlation to Caption Concreteness.} We manually annotate concreteness scores for 500 captions from LAION-400M~\cite{schuhmann2022laion}, selected to cover a wide variety of levels of concreteness. As seen in Table \ref{tab:corr}, \methodname outperforms existing methods in this setting by a large margin, demonstrating its advantage in selecting the most concrete image captions.

\begin{table}[t]
  \centering
  \setlength{\tabcolsep}{3pt}
  \begin{tabular}{lccc}
\midrule
    & $\rho$  & $\rho_s$  & $\tau$   \\
    \midrule
    Before Distillation   & 0.65 & 0.6 & 0.46 \\
    After Distillation   & 0.72 & 0.75 & 0.6 \\
    \bottomrule
  \end{tabular}
  \caption{\textbf{Distillation Effect on Caption Concreteness Correlation}. We show correlations to ground-truth annotated caption concreteness scores before and after distillation. The ``After Distillation" row corresponds to our final \methodname{} score.}
  \label{tab:dist_ablation}
\vspace{-0.1in}
\end{table}

\section{Ablations}
\label{sec:ablations}

\noindent \textbf{Distillation Concreteness Effect.}  Although the distillation procedure is necessary to make inference feasible with respect to runtime, we provide further motivation by measuring the effect of distillation on the correlation to ground-truth annotations of concreteness scores in Table \ref{tab:dist_ablation}. As can be seen, the distillation improves correlations values, providing further motivation beyond computational efficiency and simplifying the inference of our \methodname{} model. We hypothesize that this improvement is due to smoothing of noisy reconstruction of the \iae{} and \tae{} by the distillation process.

\vspace{1em}
\noindent \textbf{Distillation Speed-up.} We ablate the speed-up provided by the distillation phase by running the \tae{}, \iae{} and the distilled \methodname{} on the same hardware settings (an Nvidia A6000 GPU), the same batch size of 1 and the same caption samples. We find that the \tae{} and \iae{} process 0.45 and 0.2 samples per second respectively, and the distilled score processes 45 samples per second. Note that the time it would take to generate scores for our 8M subset of LAION-400M dataset is approximately 11,000 GPU hours for the \iae{} and 5,000 GPU hours for the \tae{} compared to just 50 GPU hours using the distilled \methodname{}. Additionally, for a batch size of 1, the distilled model takes less than 700 MB of GPU memory compared to 13GB and 14GB for the \iae{} and \tae{} respectively.

\medskip
\noindent \textbf{Use of Both \tae{} and \iae{} Scores.}
 We also ablate the use of both \tae{} and \iae{} scores for downstream captioning model training in Table \ref{tab:score_ab}. In the figure, we show captioning metrics (CIDEr and SPICE) of a model trained on a distilled version of each of the scores in isolation, compared to the combined \methodname metric which outperforms both.

\begin{table}[t]
  \centering
  \setlength{\tabcolsep}{5pt}
  \begin{tabular}{lcc|cc}
      \toprule
     & \multicolumn{2}{c}{\textbf{COCO}} & \multicolumn{2}{c}{\textbf{NoCaps}}\\
\midrule
    Method & CIDEr  & SPICE  & CIDEr  & SPICE \\
    \midrule
    \tae{}   & 17.8 & 5.9 & 15.1 & 3.3\\
    \iae{} & 29.8	& 9.4 & 27.8 &	5.8   \\
    \textbf{\methodname} & \textbf{30.9} & \textbf{9.7} & \textbf{28.2}  & \textbf{6} \\
    \bottomrule
  \end{tabular}
  \caption{\textbf{Score Ablations} We ablate the importance of using scores obtained from both the SBA and VBA pipelines over 200k samples dataset that was filtered using the different scores.}
  \label{tab:score_ab}
\vspace{-0.15in}
\end{table}

\noindent 
\textbf{\methodname{} Model Component Ablations.} In Table \ref{tab:model_parts_ablation}, we ablate the effect of various design choices in the \methodname{} pipeline by evaluating their effects on caption concreteness prediction (using the benchmark described in Section \ref{sec:concretness-eval}). In particular, we test different LLM sizes ~\cite{zhang2024tinyllama,openlm2023openllama} in the \tae{} pipeline, different captioning model architectures in the \iae{} pipeline, and the similarity measure used in each pipeline (edit distance vs. BERTScore). To identify the effect of each component, we evaluate \tae{} and \iae{} predictions in isolation (without combining or distilling them). As is seen in the table, our chosen LLM and captioning model perform comparably to the alternative models tested, showcasing the robustness of the \iae{} and \tae{} across model sizes. Moreover, while the simple edit distance similarity measure performs acceptably for the \tae{} pipeline, the BERTScore similarity measure produces significantly better correlations in the \iae{} pipeline, matching the intuition that the \iae{} is inherently lossy with respect to the precise form of texts and must rely on a more semantic measure to properly detect abstract sentences.

\begin{table}[t]
  \centering
  \setlength{\tabcolsep}{3pt}
  \begin{tabular}{llcc|ccc}
      \toprule \multicolumn{7}{c}{\textbf{Caption Concreteness}}\\
\midrule
    & Pipe & Model Part & Sim. & $\rho$  & $\rho_s$  & $\tau$   \\
    \midrule

    & \tae & TinyLLaMa-1.1B & ED & 0.59 & 0.58 & 0.45 \\
    & \tae & OpenLLaMa-3B & ED & 0.57 & 0.56 & 0.43 \\
    * & \tae & LLaMa-2-7B & ED & 0.53 & 0.51 & 0.48 \\

    & \tae & TinyLLaMa-1.1B & BSc & 0.57 & 0.56 & 0.43 \\
    & \tae & OpenLLaMa-3B & BSc & 0.56 & 0.55 & 0.42 \\
    & \tae & LLaMa-2-7B & BSc & 0.57 & 0.56 & 0.43 \\
   
    \midrule
    & \iae & BLIP-Base & ED & 0.43 & 0.4 & 0.31 \\
    & \iae & BLIP-Large & ED & 0.43 & 0.36 & 0.27 \\
    & \iae & BLIP-2  & ED & 0.44 & 0.41 & 0.31 \\
    & \iae & BLIP-Base & BSc & 0.6 & 0.6 & 0.46 \\
    & \iae & BLIP-Large & BSc & 0.58 & 0.56 & 0.43 \\
    * & \iae & BLIP-2  & BSc & 0.6 & 0.58 & 0.45 \\
    \bottomrule
  \end{tabular}
  \caption{\textbf{Ablations over \iae{} and \tae{} Design Choices.} %
  We ablate the effect of the LLM used in the \tae{} pipeline and the captioning model used in the \iae{} pipeline, as well as the text similarity measure, on the correlation to the ground-truth concreteness annotations. Note that here we measure correlation to each model of the piplines (\iae{} and \tae{}) used in isolation. BSc and ED refer to BERTScore and edit distance respectively. We report the Pearson $\rho$, Spearman $\rho_s$, and Kendall $\tau$ correlation coefficients. Our default settings are indicated with a prepended *.}
  \label{tab:model_parts_ablation}
\end{table}

\section{Related Work}

\noindent
\textbf{Evaluating Text Concreteness.} Word concreteness is a topic of interest in cognitive science~\cite{paivio1968concreteness,richardson1975concreteness,schwanenflugel2013abstract,khanna2021well}, and a number of works have studied automatic prediction of word concreteness using machine learning~\cite{hill2014multi,hill2014learning,hessel2018concreteness,rabinovich2018learning,charbonnier2019predicting,alper2023bert}. However, little attention has been paid to measuring concreteness at the caption or string level. \citet{shi2019visually} define concreteness of constituents by matching them to images for learning syntactic representations without explicit supervision; as was later shown, the signal of noun concreteness plays a key role in the model’s syntactic predictions~\cite{kojima2020learnedvisuallygroundedneural}. Most similar to us is \citet{wu2023composition}, who generate multiple images for each caption and average the CLIP similarity scores over all the images to produce a caption-level concreteness score. Other text evaluation metrics compare to reference texts~\cite{gehrmann2023repairing} or a reference image~\cite{hessel2021clipscore}, while we are interested in the inherent quality of text in isolation (namely, its visual concreteness).

\vspace{1em}
\noindent
\textbf{Multimodal Dataset Curation.}
Due to the highly noisy nature of Internet multimodal data, prior works have filtered using approaches such as rule-based text parsing~\cite{radenovic2023filtering}, using CLIP similarity to detect misaligned text-image pairs~\cite{schuhmann2022laion}, de-duplicating semantically similar content~\cite{abbas2023semdedup}, and removing samples with text that overlap with the image~\cite{maini2023t}. A number of prior works have also proposed replacing or augmenting multimodal datasets with synthetic samples~\cite{li2022blip,li2023blip,fan2023improving, lai2023scarcity, nguyen2023improving}.
By contrast, we do not require modifying the given dataset and identify semantically infelicitous captions allowed by prior methods. Our work also contrasts with dataset distillation, which has been applied to multimodal dataset curation~\cite{wu2023visionlanguage}; while dataset distillation methods select samples to explicitly optimize a chosen downstream objective, we focus on the simpler and more general task of identifying samples of inherently poor quality.

\section{Conclusion}

We present a new metric for measuring the visual concreteness of image captions without an image reference. By leveraging strong foundation models, we quantify visual-semantic information loss in an unsupervised manner and find that this highly correlates with human concreteness judgments. Our results demonstrate that \methodname{} is effective at selecting a core of high-quality image-caption samples from web-scale multimodal datasets for training models in the resource-constrained setting. We foresee the use of \methodname{} in additional tasks requiring the curation of web-scale multimodal data, where high-quality, visually-concrete text is needed.

\section*{Limitations}
While our method manages to detect visually concrete captions well, it lacks sensitivity to grammatical structure, which might cause it to label oddly phrased captions as concrete. For instance, consider the caption: ``a computer near a tree with a boy next to a table with a keyboard''. This caption is highly concrete and gets a high \methodname{} score of 1.0. However, removing all object relations from the caption produces the following: ``computer tree boy table keyboard'' which results in a relatively minor decrease of the \methodname{} score to 0.89. Such low-quality captions might have a negative impact on tasks such as image captioning where the model must learn to output grammatically correct English sentences which should ideally describe relevant fine-grained relations between entities. We hypothesize that this behavior stems from the dataset used to train the distillation model (CC3M) which is not likely to include such oddly phrased captions, and so these non-grammatical structures are not learned. We hypothesize that training over a dataset with higher caption diversity will likely alleviate this issue.

In addition, due to limited computational resources, our experiments were conducted on a relatively small scale of 8 million sample initial training dataset based on LAION-400M. We expect that increasing the scale and the filtered dataset proportionally will result in a performance improvement in the downstream model performance. However, we leave verifying this as well as testing the effect of \methodname{} filtering on other downstream tasks such as VQA and caption ranking to future work.

Finally, while our method detects and filters an important category of noise in multimodal datasets, we note that abstract captions such as those in Figure \ref{fig:teaser} may contain important information which our method discards. Future work might instead extract the relevant visual information from such captions, to avoid losing the information signal in such items. We also note that such captions often contain external or subjective information which could be of interest to tasks such as news image captioning or multimodal sentiment analysis, where external context is of interest. To identify such cases, further work might enhance the interpretability of our method to explore \emph{why} a caption is or is not concrete.

\section*{Ethics Statement}

Models trained on multimodal Internet data may inherit biases from their training data. Our method is not designed to filter potentially harmful image descriptions; moreover, such biases are also present in the models used as part of our pipeline (CLIP, generative models) and thus our model may possibly inherit or amplify these issues for downstream tasks. We anticipate further research into such biases and guidelines needed before putting these models into deployment. 

\section*{Acknowledgements}

This work was partially supported by the KLA Foundation and Google. We thank Yonatan Bitton and Keren Ganon for their helpful feedback.

\bibliography{anthology,custom}
\bibliographystyle{acl_natbib}

\clearpage
\appendices

\section{Implementations Details}

\begin{figure*}[h]
    \centering
    \begin{subfigure}[t]{0.49\textwidth}
        \centering
        \includegraphics[height=2.22in]{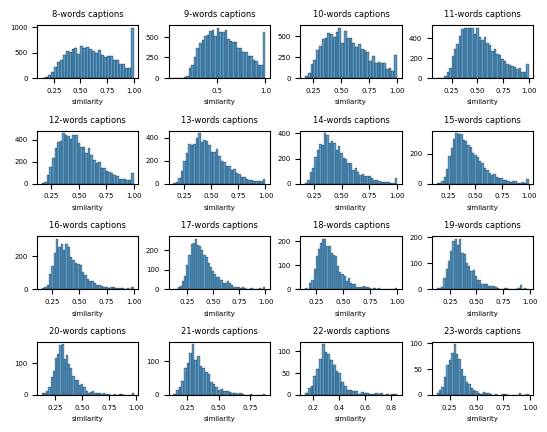}
        \caption{Before Standardization}
        \label{before}
    \end{subfigure}%
    \hspace{-1em} 
    \begin{subfigure}[t]{0.49\textwidth}
        \centering
        \includegraphics[height=2.22in]{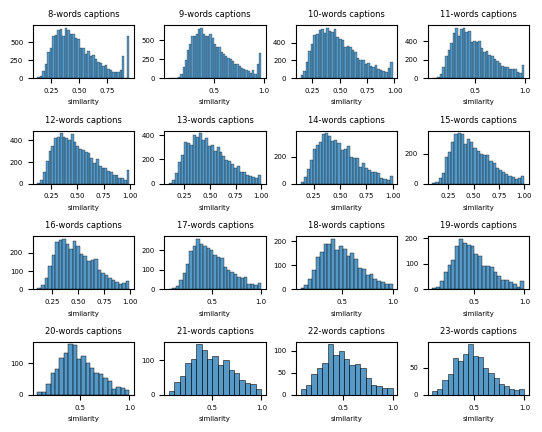}
        \caption{After Standardization}
        \label{After}
    \end{subfigure}
    \caption{\textbf{Standardizing by caption length.} We show the reconstruction similarity scores of \tae{} for each caption length before standardization (in \ref{before}) and after standardization (in \ref{After}).}
    \label{fig:caption_len}
\end{figure*}

\begin{figure}[h]
    \centering
    \includegraphics[width=1.0\linewidth]{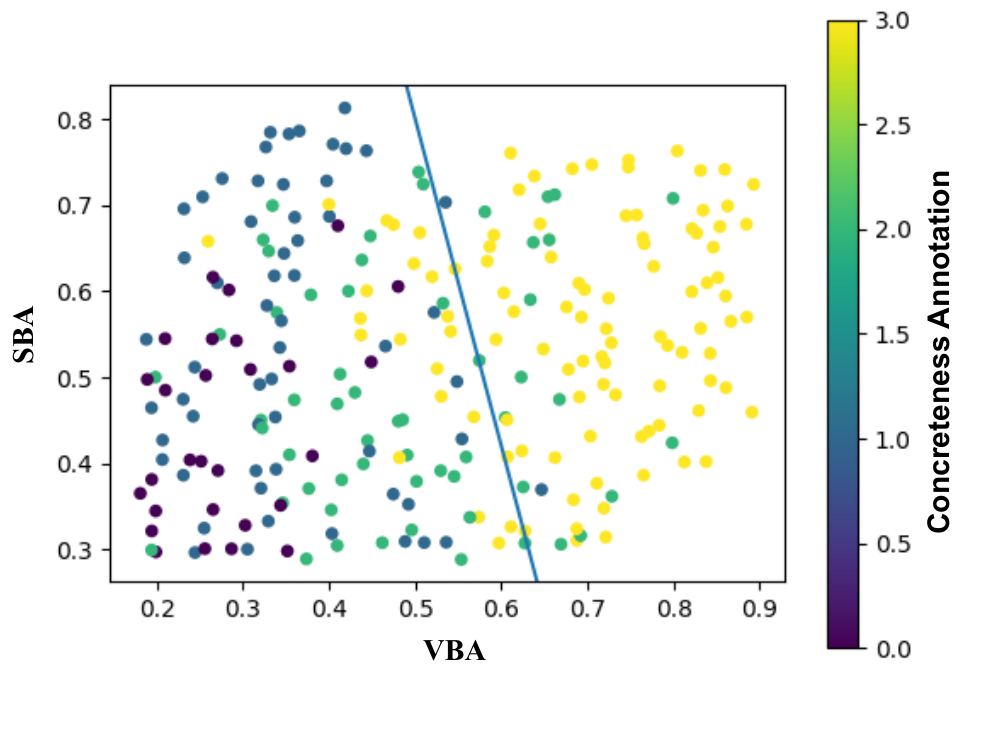}
    \vspace{-10pt}
    \caption{\textbf{Finding the Optimal Weights.} We measure the optimal combination of the two scores with respect to ground-truth concreteness annotations.} 
    \label{fig:icq_comb}
\end{figure}

\subsection{Standardizing By Caption Length}
\label{sec:caption_len}
We aim to have reconstruction scores that are only dependent on the concreteness of captions and not on the length of the captions for both the \tae{} and \iae{}. In Figure \ref{fig:caption_len}, we show the distribution of the edit-distance based reconstruction similarities of the \tae{} before and after standardization per caption length. We can see in Figure \ref{before} that there is a strong dependency on caption length, which we would like to avoid. 

More specifically, we force the reconstruction similarity distribution to be distributed according to $\mathcal{LN}(\mu=0.5, \sigma=1)$, where $\mathcal{LN}$ denotes a Logit-Normal distribution. The normalization is performed by standardizing the logit of the similarities (defined by $ln(\frac{1}{1-p})$) for each caption length, and then taking the inverse logit. We can see in Figure \ref{After} that short captions are reconstructed more easily compared to longer ones, and that normalization by caption length successfully disentangles the reconstruction scores from the caption length dependency.

\subsection{\methodname Distillation}
We distill the knowledge obtained by the two pipelines described in the paper in a two-stage manner. Firstly, we distill the \iae{} and \tae{} scores into two distinct DistilRoBERTa~\cite{liu2019roberta} models. We then collect a small subset of 244 captions, sampled to have approximately uniform joint distribution of scores, and annotate the concreteness scores of these captions. This is showcased in Figure \ref{fig:icq_comb}. We regress over these samples to get the optimal weights.

\begin{figure}[h]
    \centering
        \includegraphics[height=2.2in]{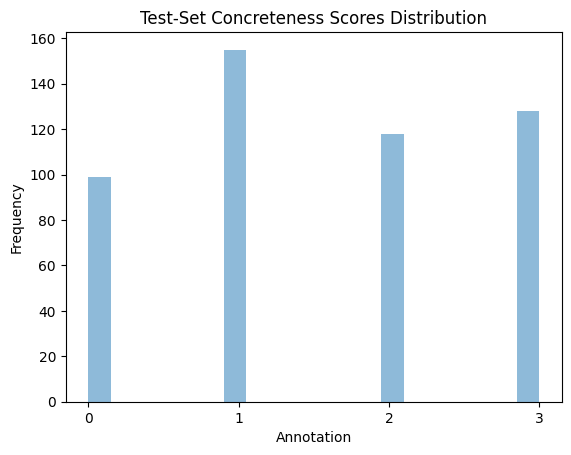}
    \vspace*{-0mm}\caption{Distribution of annotated concreteness scores in our manually labeled test set of 500 captions. All samples are from LAION-400M. Annotations range from highly abstract (0) to highly concrete (3).} 
     \label{fig:dist_ann}
\end{figure}

\subsection{Caption Concreteness Benchmark Distribution}
Our aim is to have a small, yet diverse set of samples that represent the wide diversity of possible captions. Since Laion-400M is very noisy and only a small portion of it includes highly
concrete captions, we curate our captions to achieve a balanced distribution of concreteness scores, as illustrated in Figure \ref{fig:dist_ann}. As seen there, the concreteness of the benchmark's captions is evenly distributed between abstract and concrete concepts.

\subsection{Zero-Shot CLIP Concreteness Score}
We adapt the Stroop Probing method~\cite{alper2023bert} for estimating text concreteness. While \citet{alper2023bert} test this on single words, we adapt this method to captions by replacing the empty slot in prompts with a caption rather than a single word. We use their prompts, omitting those which do not match the context of an entire caption being inserted in the masked slot (i.e., omitting the prompts ``Alice giving the [*] to Bob'' and ``Bob giving the [*] to Alice'').

\subsection{GPT Prompts}
The following prompts were used to extract concreteness scores for image captions\footnote{To get the concreteness scores of words, we used the same prompts with ``word'' instead of ``caption'' in the appropriate places.} from GPT-3.5 and GPT-4o:

\noindent\texttt{\textbf{System}: ``You are an expert visual reasoner, capable of understanding the visual concretess of image captions. A visually concrete caption is a caption that is highly visual, and can be vividly imagined.''}

\noindent \texttt{\textbf{User}: ``Provide a numerical score on a scale of 1-5, when 1 is non-visual and 10 is highly visual caption for the following caption : <caption>. Only provide the numerical score and nothing else.''}

Note that we experimented with three different ranges of [1-N] of concreteness scores in our prompts: N=3, N=5 and N=10.  We found that N=5 yielded the best results.

\subsection{aveCLIP Word Concreteness}
Since aveCLIP requires generating many images per word or caption, we found that running aveCLIP over the entire word concreteness dataset is not feasible due to runtime constraints. Therefore, we evaluate its performance on a random subset of 150 words/captions.

\subsection{Training Hyperparameters and Additional Information}
\label{subsec:additional_info}
\vspace{0.5em}
\textbf{\tae{}.} We train the linear layer of the \tae{} with a batch-size of 128, learning rate of 2e-3 with cosine scheduler and a warm-up ratio of 0.03, and train for a two epoch over a single Nvidia-A6000 GPU. All other hyperparameters are set to the defaults of the HuggingFace Trainer API.

\vspace{0.5em}
\noindent\textbf{\iae{} Text-To-Image.} For the image generation of the diffusion model in the \iae{}, we use guidance scale of 9 and 20 inference steps.

\subsection{Model Checkpoints Used}
We detail here all the checkpoints that were used in our experiments. All model checkpoints are taken from the Hugging Face Model Hub\footnote{\url{https://www.huggingface.co/models}}.
For the \tae{}, we used:
\begin{itemize}[noitemsep]

    \item \texttt{openai/clip-vit-large-patch14} (only the text encoder)
    \item \texttt{meta-llama/Llama-2-7b}
\end{itemize}
For the \iae{}, we used:
\begin{itemize}[noitemsep]
    \item \texttt{stabilityai/stable-diffusion-2}
    \item \texttt{Salesforce/blip2-opt-2.7b}
\end{itemize}
For the distilled model, we used:
\begin{itemize}[noitemsep]
    \item \texttt{distilroberta-base}
\end{itemize}
For training a captioning model, we used:
\begin{itemize}[noitemsep]
    \item \texttt{microsoft/swin-base-patch4\\-window7-224-in22k}
    \item \texttt{gpt2}
\end{itemize}
For training a dual-encoder model, we used:
\begin{itemize}[noitemsep]
    \item \texttt{bert-base-uncased}
    \item \texttt{google/vit-base-patch16-224}
\end{itemize}

\subsection{Finding the Score Combination Parameters}
To compute the combination parameters of the \tae{} and \iae{} scores, we label 244 captions, sampled uniformly over \iae{} and \tae{} scores, with concreteness scores in the range $0$--$3$. We use logistic regression to find the parameters $a,b,c$ of  $\sigma(a \cdot \iae{} + b \cdot \tae{} + c)$, where $\sigma(x)=\frac{1}{1 + e^{-x}}$ is the sigmoid function, such that the output will approach 1 for concrete captions and 0 for abstract ones. We label concrete captions as captions with concreteness above the median score in the labeled dataset and abstract captions as captions with a score below this median. We visualize the annotated samples and the regression line  $a \cdot \iae{} + b \cdot \tae{} + c = 0$ in Figure \ref{fig:icq_comb}. The parameters found and used in our \methodname are $a = 13.2$, $b = 3.6$ and $c = -9.4$. As seen in the figure, both scores contribute to the optimal predicted concreteness score, validating the importance of using both SBA and VBA components together in our full pipeline.

\section{Additional Qualitative Examples}

We visually show examples of each of the scores' weaknesses and the way they compliment each other. In Figure \ref{fig:vba_sba_conc}, we show examples of \emph{concrete} captions, the reconstructed captions by \iae{} and \tae{}, and the different scores of each of them. The first four rows exemplify why \iae{} may fail to reconstruct some concrete captions. For instance, the caption ``a nurse mopping a surgeon's brow during an operation in an operation pub'' was reconstructed to 
``two people in protective gear'' which bears relatively low semantic similarity to the original caption. These cases mainly stem from the inherent difficulty of reconstructing (through a captioning model) from an image the exact caption from which the image was generated, as there may be many possible such captions. In this case, the use of \tae{} helps determining that the caption is concrete. 

In a complementary manner, we show in Figure \ref{fig:vba_sba_abs} examples of \emph{abstract} captions. In this figure, the first four rows demonstrate that using \tae{} alone is also not enough, as it is sometimes able to reconstruct abstract captions due to the higher semantic information that is contained in the CLIP embeddings. In this scenario, \iae{} compensates for these failures, as it is very unlikely to reconstruct abstract text.

These qualitative examples further illustrate the benefit of using both \iae{} and \tae{}. Indeed, in Figures \ref{fig:vba_sba_conc}--\ref{fig:vba_sba_abs}, it can be observed that \methodname{} reflects the advantages of both pipelines by generating low scores for abstract captions, and high scores for concrete ones in a consistent manner.

\label{sec:appendix_ablations}

\begin{figure*}[h]

\centering
\begin{adjustbox}{width=\textwidth,center} 
\begin{tabular}{  p{3.4cm} | p{3.2cm} | p{2.4cm} | p{1.8cm}| p{0.6cm} |p{0.6cm} | p{0.6cm} }
        \toprule
\textbf{Input caption}    
& \textbf{\tae{} reconstructed caption}   
& \textbf{\iae{} reconstructed caption} & \textbf{\iae{} bottleneck image} & \textbf{\tae{}} & \textbf{\iae{}} &\textbf{\methodname} \\\midrule

a nurse mopping a surgeon's brow during an operation in an operation pub	
& a nurse wiping the brow of a surgeon during an operation in an operating room
&two people in protective gear	
& \includegraphics[scale=0.12,valign=t]{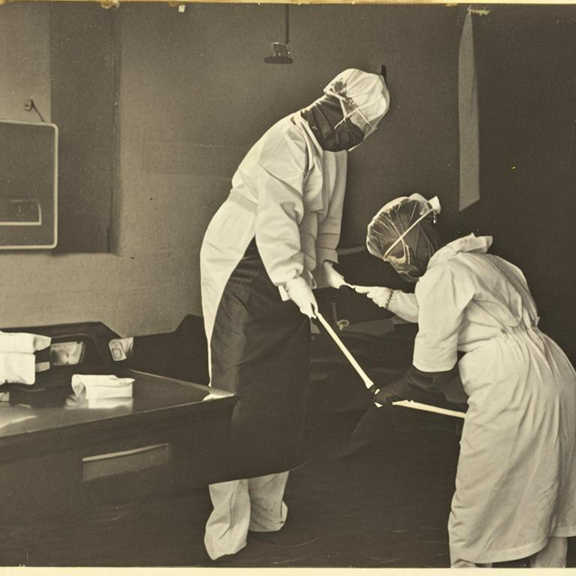} 
& 0.77
& \hlcr{0.25}
& 0.72
\\\hline

bougainvillea climbing up the wall of a villa
& bougainvillea climbing on a wall of a villa
& a house covered in pink flowers	
& \includegraphics[scale=0.12,valign=t]{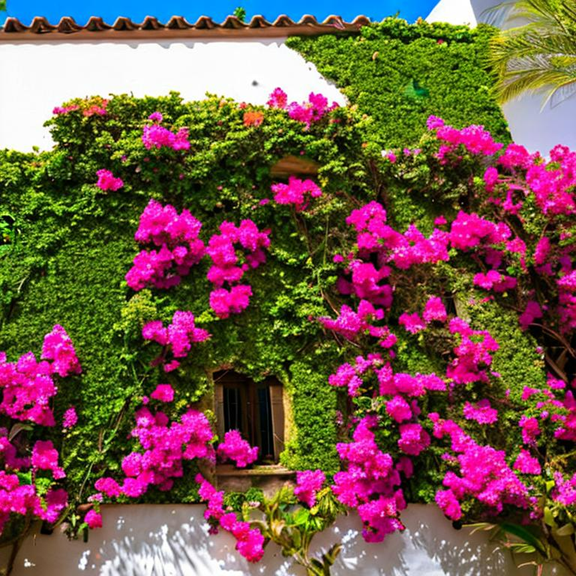} 
& 0.72
& \hlcr{0.26}
& 0.81
\\\hline

table top shot of many vegetables and mexican bugs on a table 
& close up shot of vegetables and bugs on a table
& vegetables arranged in the shape of a human head 
& \includegraphics[scale=0.12,valign=t]{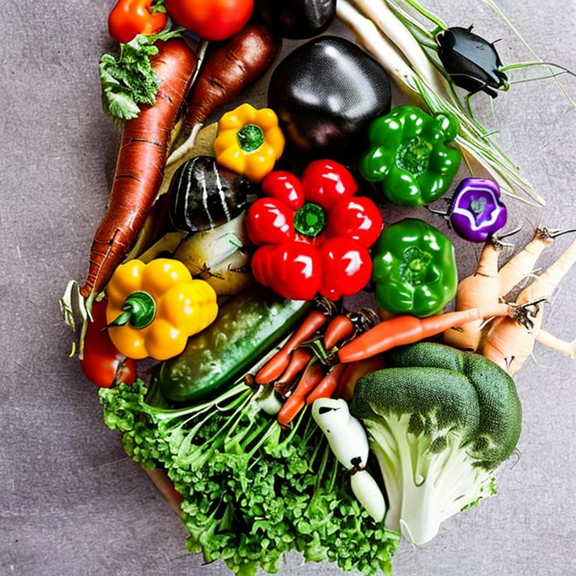} 
& 0.70
& \hlcr{0.25}
& 0.76

\\\hline

silhouette of a man with a gun in poses royalty
& silhouette of a man holding a gun in poses royalty
& a group of people silhouettes on a white background
& \includegraphics[scale=0.12,valign=t]{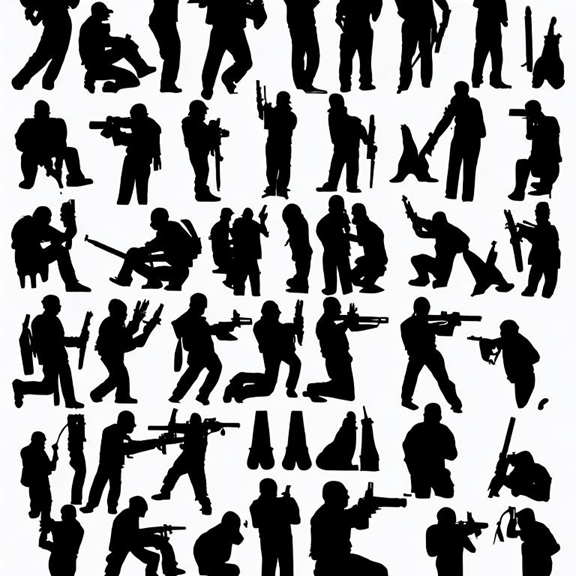} 
& 0.82
& \hlcr{0.26}
& 0.93

\\\hline

small flock of sheep in winter snow on a hilltop
& small flock of sheep in snow on a hill	
& a herd of sheep in the snow
& \includegraphics[scale=0.12,valign=t]{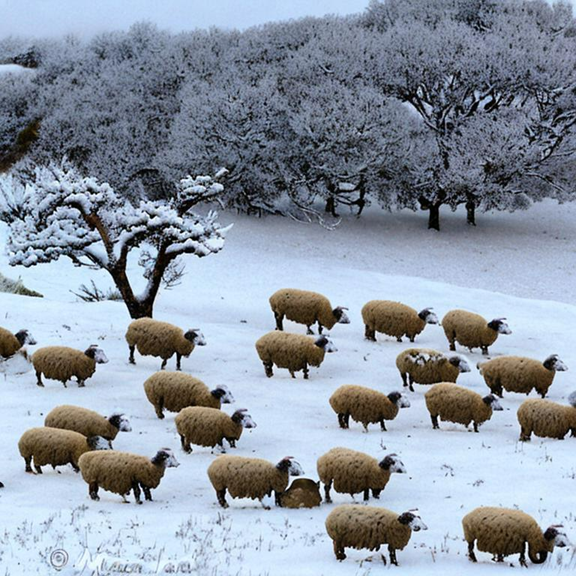} 
& 0.72
& 0.95
& 1.0
\\\hline

small blue and white airplane parked on the ramp with a control tower in the distance
& small blue and white airplane parked on the tarmac next to a control tower
& a blue and white airplane parked on the tarmac
& \includegraphics[scale=0.12,valign=t]{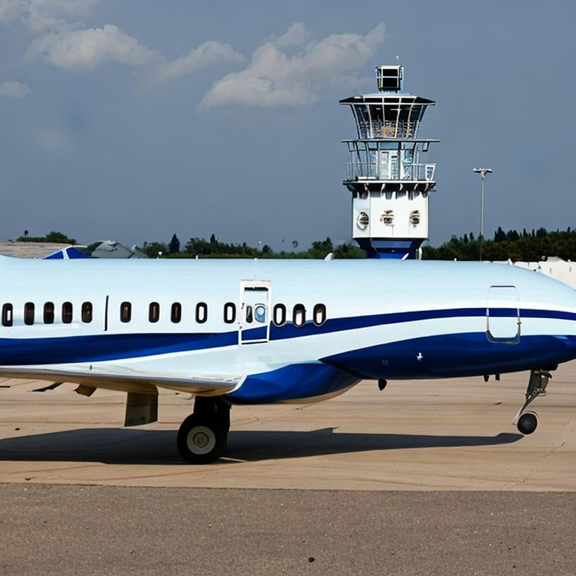} 
& 0.96
& 0.95
& 1.0

\\\hline

a young girl runs through a field of cabbages
& a young girl runs through a field of cabbages
& a girl walking through a field of cabbage
& \includegraphics[scale=0.12,valign=t]{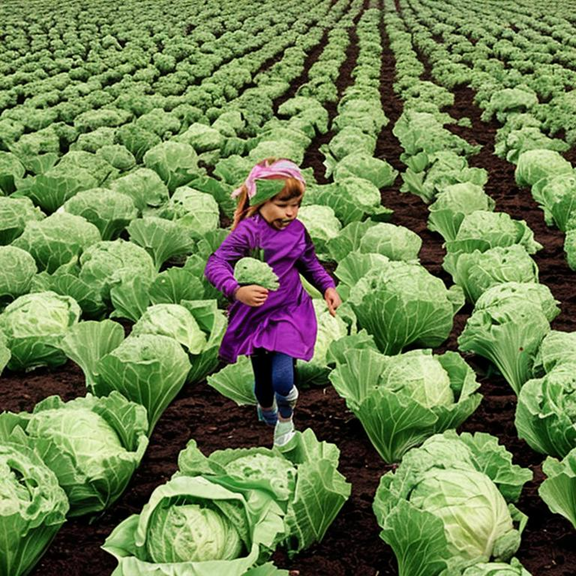} 
& 0.96
& 0.95
& 1.0

\\\hline
a red post box and a telephone box stand together in a village
&a red telephone box and a post box stand together in a village
& a red post box next to a stone wall
& \includegraphics[scale=0.12,valign=t]{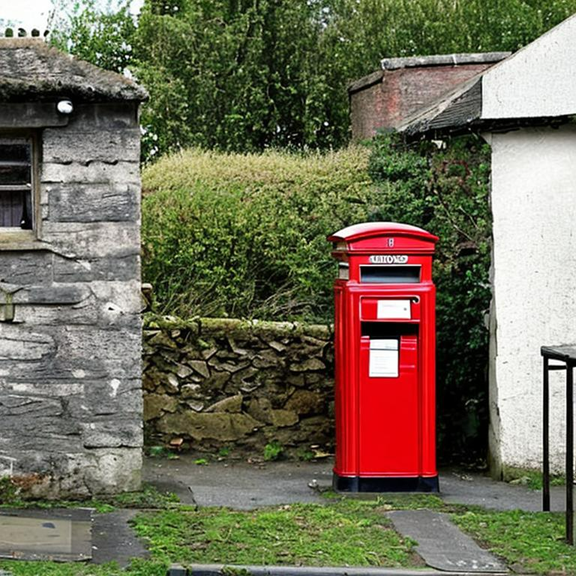} 
& 0.84
& 0.89
& 0.92

\\\bottomrule
      
\end{tabular}
\end{adjustbox}

\caption{\textbf{Qualitative Examples for Highly Concrete Captions}. We demonstrate reconstructions of highly concrete captions and the final distilled \methodname scores. We mark by \hlcr{red} low reconstruction scores which correspond to unsuccesfull detection of the concrete captions. As illustrated above, VBA yields generally less consistent scores for concrete captions (see the text for further discussion). Nonetheless, our final distilled scores correctly identify these captions as concrete ones, obtaining high \methodname{} scores over these captions.}

\label{fig:vba_sba_conc}

\end{figure*}

\begin{figure*}[h]

\centering
\begin{adjustbox}{width=\textwidth,center} 
\begin{tabular}{  p{3.5cm} | p{4.2cm} | p{2.2cm} | p{1.8cm}| p{0.6cm} |p{0.6cm} | p{0.6cm} }
        \toprule
\textbf{Input caption}    
& \textbf{\tae{} reconstructed caption}   
& \textbf{\iae{} reconstructed caption} & \textbf{\iae{} bottleneck image} & \textbf{\tae{}} & \textbf{\iae{}} &\textbf{\methodname} \\\midrule

 keep an eye on the ball when it comes to investments
 &keep an eye on the ball when it comes to investments	
& a soccer ball on a green field
& \includegraphics[scale=0.12,valign=t]{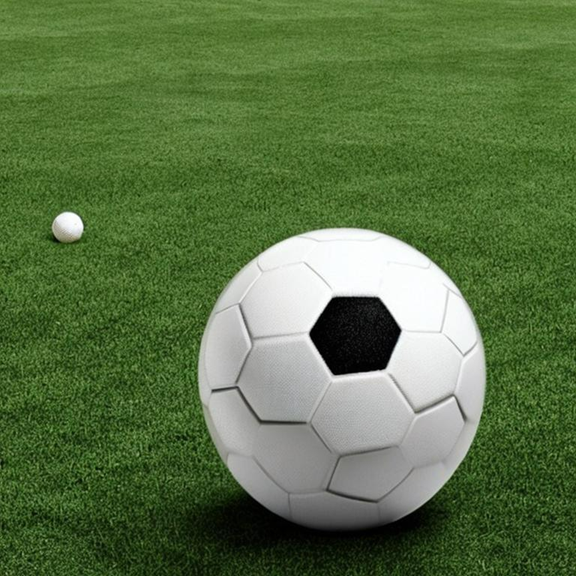} 
& \hlcr{0.91}
& 0.19
& 0.1

\\\hline

what 's the best thing about having a best friend of the opposite gender ?	
& the best thing about having a friend of the opposite gender	
& two young women sitting on a bench 
& \includegraphics[scale=0.12,valign=t]{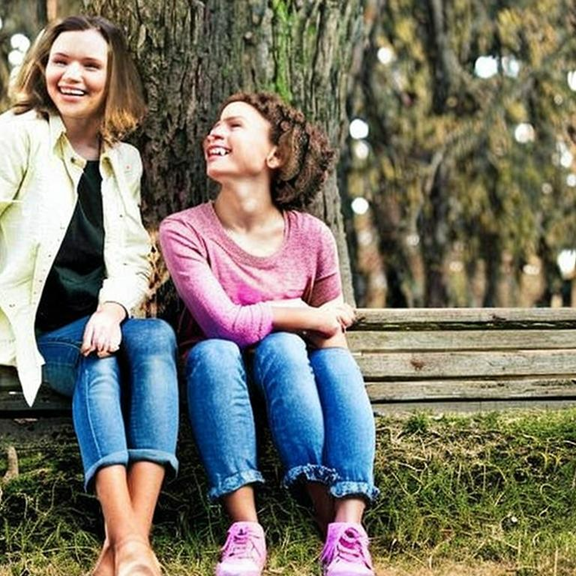} 
& \hlcr{0.89}
& 0.16
& 0.1

\\\hline

film character : would you like to bet on these shares this christmas ?
& which film character would you like to see in your shares this christmas?
&santa claus, santa claus and sant
& \includegraphics[scale=0.12,valign=t]{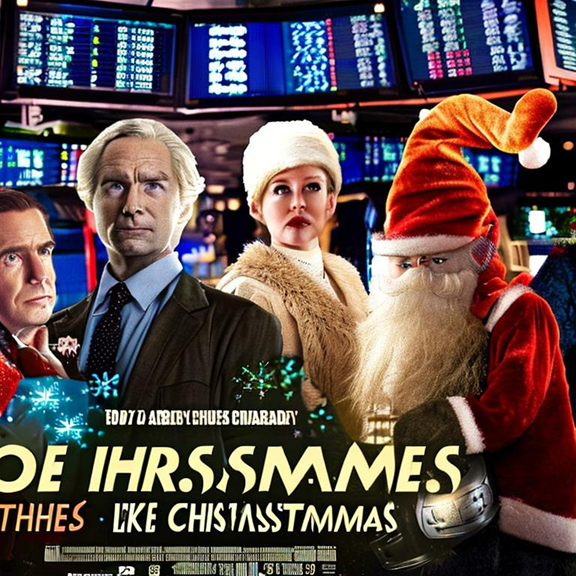} 
& \hlcr{0.79}
& 0.1
& 0
\\\hline
this is located in my home town !
& this is located in my hometown!
& a sign in front of a statue	
& \includegraphics[scale=0.12,valign=t]{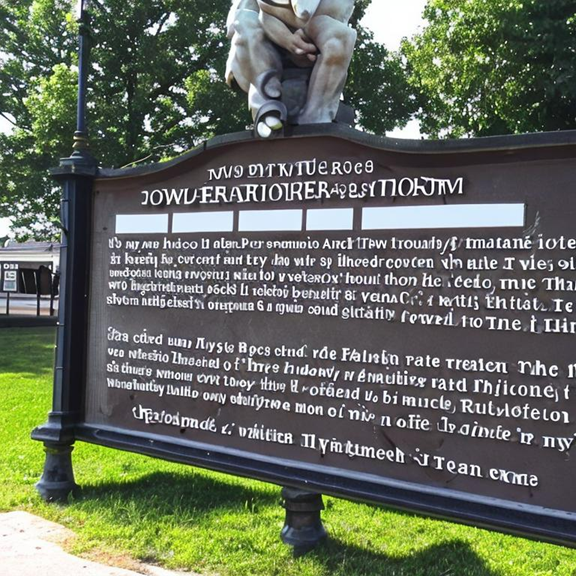} 
& \hlcr{0.75}
& 0.28
& 0

\\\hline
chaotic systems are sometimes described using fractal patterns
& fractals are patterns that can be found in many forms, such as chaotic systems and natural structures.
& a black and white tunnel	
& \includegraphics[scale=0.12,valign=t]{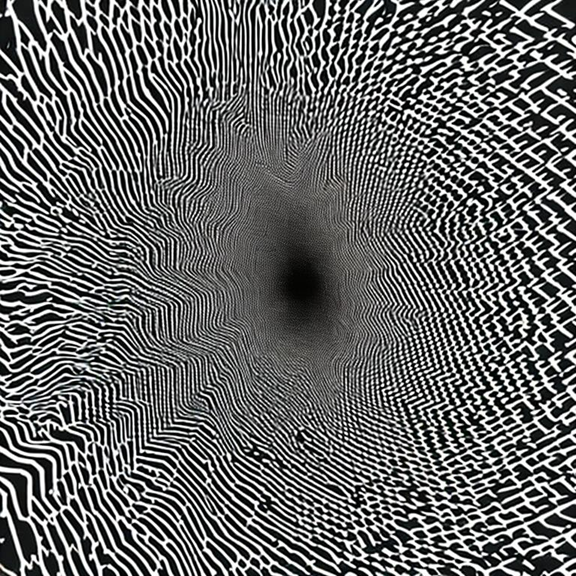} 
& 0.22
& 0.19
& 0

\\\hline

on an average , the sloth travels feet a day
 &a sloth spends most of the day on its feet
& a sloth hanging from a branch	
& \includegraphics[scale=0.12,valign=t]{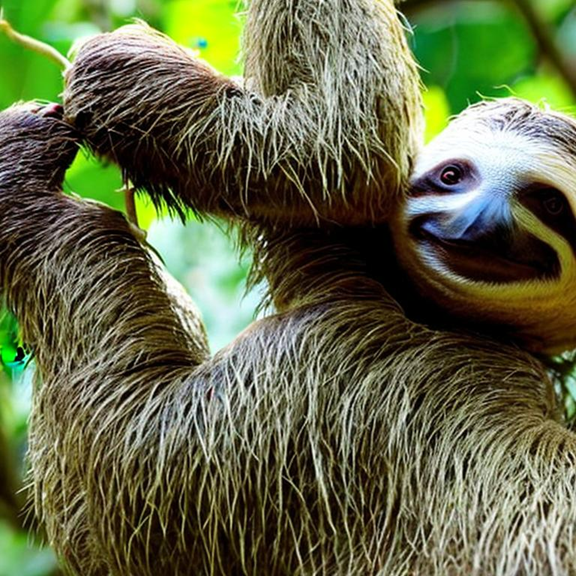} 
& 0.17
& 0.27
& 0

\\\hline
 get tips for biological genus , more commonly known as air plants , in your home
 &learn how to care for air plants, one of
& a bunch of air plants on a brown surface
& \includegraphics[scale=0.12,valign=t]{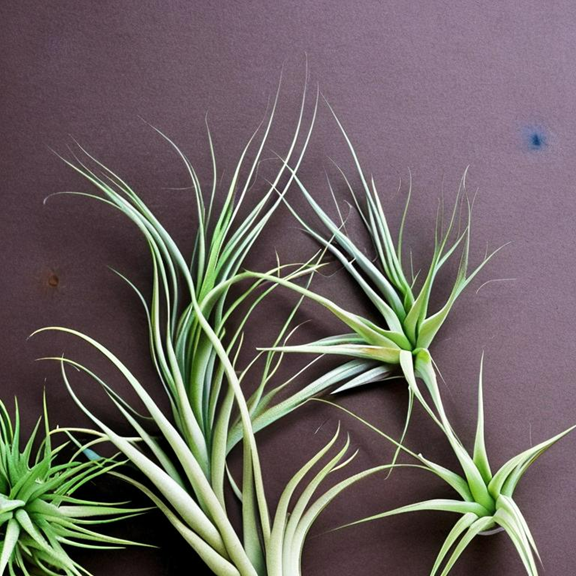} 
& 0.32
& 0.25 
& 0

\\\hline
versatile and highly capable , there 's more to this tiny camera than its giant zoom
 &this little camera packs a big punch with its zoom lens and 2
& a camera on a wooden table
& \includegraphics[scale=0.12,valign=t]{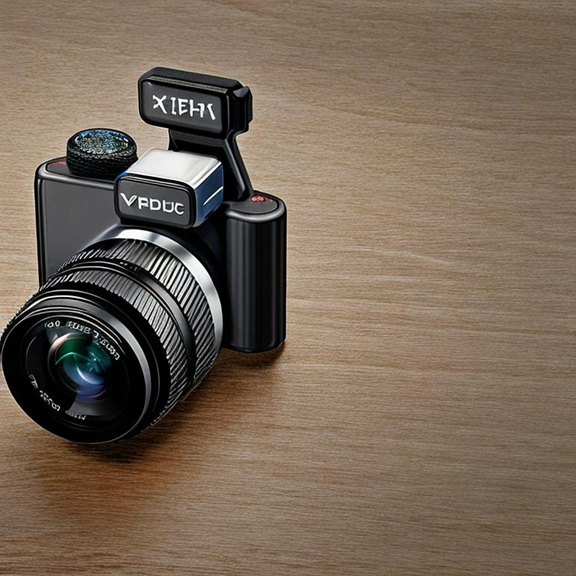} 
& 0.25
& 0.24
& 0

\\\bottomrule

\end{tabular}
\end{adjustbox}

\caption{\textbf{Qualitative Examples for Highly Abstract Captions}. We demonstrate reconstructions of highly abstract captions and the final distilled \methodname scores. We mark by \hlcr{red} captions which were reconstructed well (note that in the case of abstract captions, high scores correspond to unsuccessful detections of the abstract captions). As illustrated above, SBA yields generally less consistent scores for abstract captions (see the text for further discussion). Nonetheless, our final distilled scores correctly identify these captions as abstract ones, obtaining low \methodname{} scores over these captions. }

\label{fig:vba_sba_abs}

\end{figure*}

\end{document}